\documentclass[runningheads]{llncs}
\usepackage{graphicx}

\usepackage{booktabs}
\usepackage{color}
\usepackage{amssymb}
\usepackage{amsmath}
\usepackage{amssymb}
\usepackage{mathrsfs}
\usepackage{eufrak}
\usepackage{eucal}
\usepackage{multirow}
\usepackage{booktabs}

\usepackage{amsfonts}
\usepackage{bm}
\usepackage{import}
\usepackage{graphicx}

\usepackage{float}
\usepackage{makecell}
\usepackage[title]{appendix}
\usepackage{cite}

\begin{document}

\title{E2TIMT: Efficient and Effective Modal Adapter for Text Image Machine Translation}
\titlerunning{Efficient and Effective Modal Adapter for Text Image Machine Translation}

\author{Cong Ma\inst{1,2} \and
Yaping Zhang\inst{1,2}\thanks{Corresponding author.} \and
Mei Tu\inst{4} \and Yang Zhao\inst{1,2} \and Yu Zhou\inst{2,3} \and Chengqing Zong\inst{1,2}}
\authorrunning{C. Ma et al.}

\institute{School of Artificial Intelligence, University of Chinese Academy of Sciences, \\Beijing 100049, P.R. China \and State Key Laboratory of Multimodal Artificial Intelligence Systems (MAIS), Institute of Automation, Chinese Academy of Sciences, Beijing 100190, P.R. China \and Fanyu AI Laboratory, Zhongke Fanyu Technology Co., Ltd, \\Beijing 100190, P.R. China \and Samsung Research China - Beijing (SRC-B)\\
\email{\{cong.ma, yaping.zhang, yang.zhao, yzhou, cqzong\}@nlpr.ia.ac.cn, mei.tu@samsung.com}}

\maketitle              
\vspace{-0.35cm}
\begin{abstract}
Text image machine translation (TIMT) aims to translate texts embedded in images from one source language to another target language.
Existing methods, both two-stage cascade and one-stage end-to-end architectures, suffer from different issues.
The cascade models can benefit from the large-scale optical character recognition (OCR) and MT datasets but the two-stage architecture is redundant.
The end-to-end models are efficient but suffer from training data deficiency.
%
To this end, in our paper, we propose an end-to-end TIMT model fully making use of the knowledge from existing OCR and MT datasets to pursue both an effective and efficient framework.
More specifically, we build a novel modal adapter effectively bridging the OCR encoder and MT decoder.
End-to-end TIMT loss and cross-modal contrastive loss are utilized jointly to align the feature distribution of the OCR and MT tasks.
Extensive experiments show that the proposed method outperforms the existing two-stage cascade models and one-stage end-to-end models with a lighter and faster architecture. Furthermore, the ablation studies verify the generalization of our method, where the proposed modal adapter is effective to bridge various OCR and MT models.~\footnote{Our codes are available at: https://github.com/EriCongMa/E2TIMT}


\end{abstract}

\section{Introduction}
Text image machine translation (TIMT) is the core research of many applications, such as scene text translation, document image translation, and photo translation.
Approaches to TIMT are mainly divided into two categories: two-stage cascade method
~\cite{manga_translation, Shekar2021OpticalCR, DBLP:conf/lt4dh/AfliW16, DBLP:journals/ijdar/ChenCN15, DBLP:conf/icdar/DuHSS11}
and one-stage end-to-end method~\cite{DBLP:conf/icdar/SuLZ21,DBLP:conf/icpr/ChenYZYL20,mansimov-etal-2020-towards}. 
The cascade model deploys recognition and translation models sequentially, which benefits from training with existing large-scale optical character recognition (OCR) and machine translation (MT) datasets.
However, the task gap between OCR and MT models might hurt the performance because translation models are vulnerable to recognition errors.
Furthermore, the cascade model is two-stage, \textit{i.e.}  the sequential integration of OCR and MT models, thus is redundant in parameters and has a slow decoding speed.
To alleviate the error propagation problem, some studies turn to exploring one-stage end-to-end architecture with fewer parameters and faster decoding speed~\cite{mansimov-etal-2020-towards}. However, the scarcity of end-to-end TIMT data limits the performance of end-to-end models.
Although the multi-task learning enhanced end-to-end TIMT model incorporates external OCR datasets~\cite{DBLP:conf/icdar/SuLZ21,DBLP:conf/icpr/ChenYZYL20} or MT datasets~\cite{DBLP:conf/icpr/MaZTHWZ022}, 
the huge potential of fully benefiting from the knowledge of existing OCR and MT datasets or their corresponding pre-trained models is seldom explored. RTNet~\cite{DBLP:conf/icdar/SuLZ21} is proposed to link the OCR encoder and MT decoder, but it ignores the task gap between recognition and translation tasks, causing limited performance.
In summary, the following three major challenges are usually faced in the TIMT study:
\begin{itemize}
\item[$\bullet$] \textbf{Task Gap}. 
There is a large domain gap between the OCR/MT tasks, which indicates the direct connection of the recognition and translation models is not optimal.
\item[$\bullet$] \textbf{Cascade Redundancy}. 
It leads to model/complexity redundancy when directly cascading existing OCR and MT models without any optimization.
\item[$\bullet$] \textbf{End-to-end Data Scarcity}. 
The dataset for end-to-end TIMT is scarce. It is critical to transfer knowledge from existing OCR and MT datasets or pre-trained models, which is seldom explored by previous methods.
\end{itemize}

In this paper, we propose a novel modal adapter architecture to improve the end-to-end TIMT model by eliminating task gaps and making full of the knowledge from pre-trained OCR and MT models. Furthermore, the modal adapter can is a parameter efficient fine-tuning method, which just optimizes parameters of modal adapter by frozen pre-trained encoder and decoder. Thus, modal adapter based TIMT model has much fewer parameters to update compared with end-to-end models and has a faster inference speed than cascade models.
In detail, a self-attention based modal adapter is incorporated between the pre-trained OCR encoder and MT decoder. 
Different from vanilla adapter tuning~\cite{DBLP:conf/nips/RebuffiBV17}, which is just fine-tuned on downstream tasks, 
the task gap is bridged in our framework by a cross-modal contrastive loss that aligns the distributions between the OCR and MT features of the same sentence content. Two types of modal adapters are studied to validate the effectiveness of bridging various OCR and MT modules. Embedding modal adapter (EmbMA) is proposed to bridge OCR image encoder and MT sequential encoder, while sequential modal adapter (SeqMA) is inserted between OCR Sequential encoder and MT decoder.
Finally, the MT decoder generates the translation from the features transformed by the modal adapter.
Our contributions are summarized as follows:

\begin{itemize}
\item[$\bullet$] We propose a modal adapter based TIMT model to unify cascade and end-to-end models by bridging the pre-trained recognition encoder and translation decoder.
\item[$\bullet$] Cross-modal contrastive learning is incorporated to align the distribution of image features and text features encoded by an OCR encoder and an MT encoder respectively, which alleviates the OCR-MT task gap and improves the performance of text image machine translation.
\item[$\bullet$] Extensive experiments show our method outperforms both the existing cascade models and end-to-end models with a lighter and faster architecture. Furthermore, the modal adapter has a good generalization when bridging various recognition encoders and translation decoders.
\end{itemize}
\vspace{-0.35cm}

\begin{figure*}[t]
	\centering
	\includegraphics[scale=0.45]{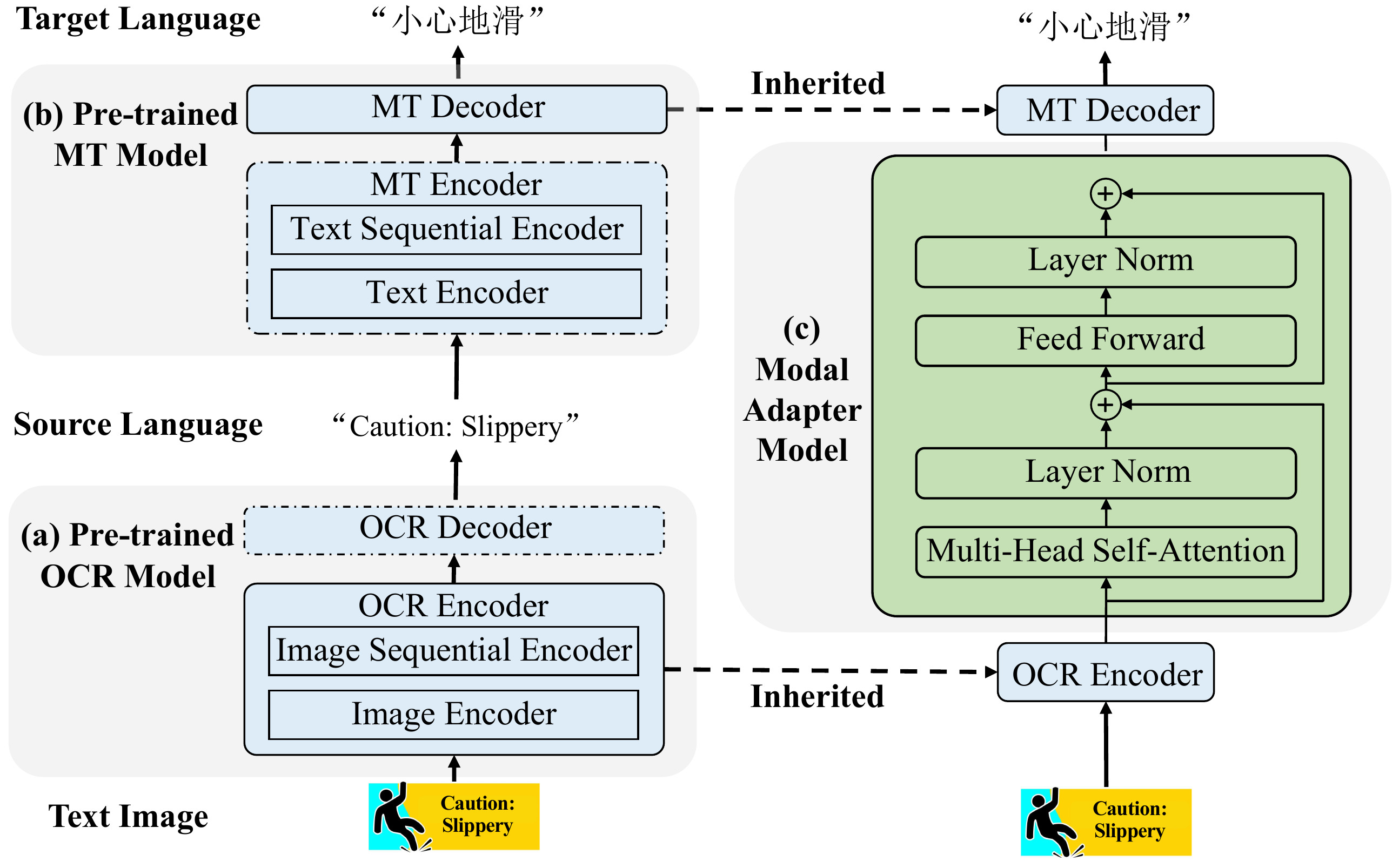}
	\caption{Architectures of OCR, MT, and Modal Adapter for TIMT Model. The solid arrow lines represent the data flow in the model. The dotted arrow lines denote the parameters of encoder and decoder in modal adapter based TIMT are inherited from pre-trained OCR/MT models.}
	\label{fig_overall_workflow}       
	\vspace{-0.45cm}
\end{figure*}

\section{Preliminary}
To unify the processing progress of recognition and translation models, we divide both the OCR and MT encoders into two submodules: image/text encoder for embedding encoding, and sequential encoder for contextual feature extraction.
We will introduce the processing flow of OCR and MT models individually.

\subsection{OCR Model}
As shown in Figure~\ref{fig_overall_workflow} (a), given an input image $I$, a  convolutional neural network (CNN) based image encoder extracts the image embedding $E_I$ by transforming image pixels into feature vectors:

\begin{equation}
E_I = \text{CNN}(I)
\end{equation}
where $I \in \mathbb{R}^{H\times W\times C}$ and $E_I \in \mathbb{R}^{l_I\times c}$. $H$, $W$, and $C$ denote the height, width, and channel of the input image respectively. $l_I$ represents the length of image embedding, which is calculated as $l_I=h\times w$, where $h$, $w$, and $c$ denote the height, width, and channel of the encoded feature map separately.

The image encoder mainly extracts the local features of the input images, while the image sequential encoder aims to model contextual information by considering the whole input sequence:

\begin{equation}
S_I = \text{Seq}_I(E_I)
\end{equation}
where $\text{Seq}_{I}(\cdot)$ represents the image sequential encoder and transformer encoder~\cite{DBLP:conf/nips/VaswaniSPUJGKP17} is utilized in our implementation. $S_I\in \mathbb{R}^{l_S\times d_S}$ denotes the sequential features in OCR model. $l_S$ and $d_S$ denote the length and dimension of sequential features respectively.

Finally, the OCR decoder generates recognized tokens auto-regressively given sequential features:

\begin{equation}
\begin{aligned}
D_I &= \text{Dec}_I(S_I);\\
P(X|I)&=\text{Softmax}(W_ID_I)
\end{aligned}
\end{equation}
where $\text{Dec}_I(\cdot)$ represents the OCR decoder, and transformer decoder~\cite{DBLP:conf/nips/VaswaniSPUJGKP17} is utilized in our implementation. $D_I$ denotes the outputs of the decoder. $W_I \in \mathbb{R}^{|\mathcal{V}_{X}|\times d_I}$ represents the linear transformation that maps the decoder features into corresponding recognized tokens, $\mathcal{V}_{X}$ is the recognition vocabulary, and $d_I$ is the dimension of decoder hidden states.
\vspace{-0.35cm}
\subsection{MT Model}
MT model translates the source language into the target language as shown in Figure~\ref{fig_overall_workflow} (b). Given a source language sentence $T$, the text encoder first maps the input words into a sequence of word embeddings:

\begin{equation}
E_T = \text{Embedding}(T)
\end{equation}
where $E_T\in \mathbb{R}^{l_E\times d_E}$ denotes the text embedding. $l_E$ and $d_E$ represent the sequence length and the dimension of text embedding respectively.

Text sequential encoder further extracts contextual features based on text embeddings:

\begin{equation}
S_T = \text{Seq}_T(E_T)
\end{equation}
where $\text{Seq}_T(\cdot)$ represents the text sequential encoder, which is a transformer encoder in our implementation. 
$S_T$ denotes the encoded text sequential features.

MT decoder finally generates the target tokens auto-regressively given sequential features:

\begin{equation}
\begin{aligned}
D_T &= \text{Dec}_T(S_T);\\
P(Y|T)&=\text{Softmax}(W_TD_T)
\end{aligned}
\end{equation}
where $\text{Dec}_T(\cdot)$ represents the MT decoder, and the transformer decoder is utilized in our implementation. $D_T$ is the output of the decoder and $W_T\in\mathbb{R}^{|\mathcal{V}_{Y}|\times d_T}$ is the linear transformation. ${\mathcal{V}_{Y}}$ represents the target language vocabulary and $d_T$ denotes the dimension of the hidden states.

\section{Methodology}

To bridge the pre-trained OCR encoder and the MT decoder, the modal adapter is proposed to transform the OCR features into the MT feature space as shown in Figure~\ref{fig_overall_workflow} (c). 

Specifically, features of the OCR encoder are transformed by the stacked modal adapter layer:

\begin{equation}
\begin{aligned}
\hat{H}_\text{MA}^{n}&=\text{LN}(\text{MSA}(H_\text{MA}^{n-1}))+H_\text{MA}^{n-1}\\
{H}_\text{MA}^{n}&=\text{LN}(\text{FFN}(\hat{H}_\text{MA}^{n}))+\hat{H}_\text{MA}^{n}\\
\end{aligned}
\end{equation}
where $H_\text{MA}^{n}$ denotes the output of the $n$-th modal adapter layer, and $H_\text{MA}^{0}$ is the feature from the OCR encoder. 
$\text{MSA}(\cdot)$, $\text{FFN}(\cdot)$, and $\text{LN}(\cdot)$ represent multi-head self-attention, feed-forward, and layer norm modules respectively. After transformation by the modal adapter, features encoded by the OCR encoder are further fed into the MT decoder to generate translation results.

\begin{figure*}[t]
	\centering
	\includegraphics[scale=0.36]{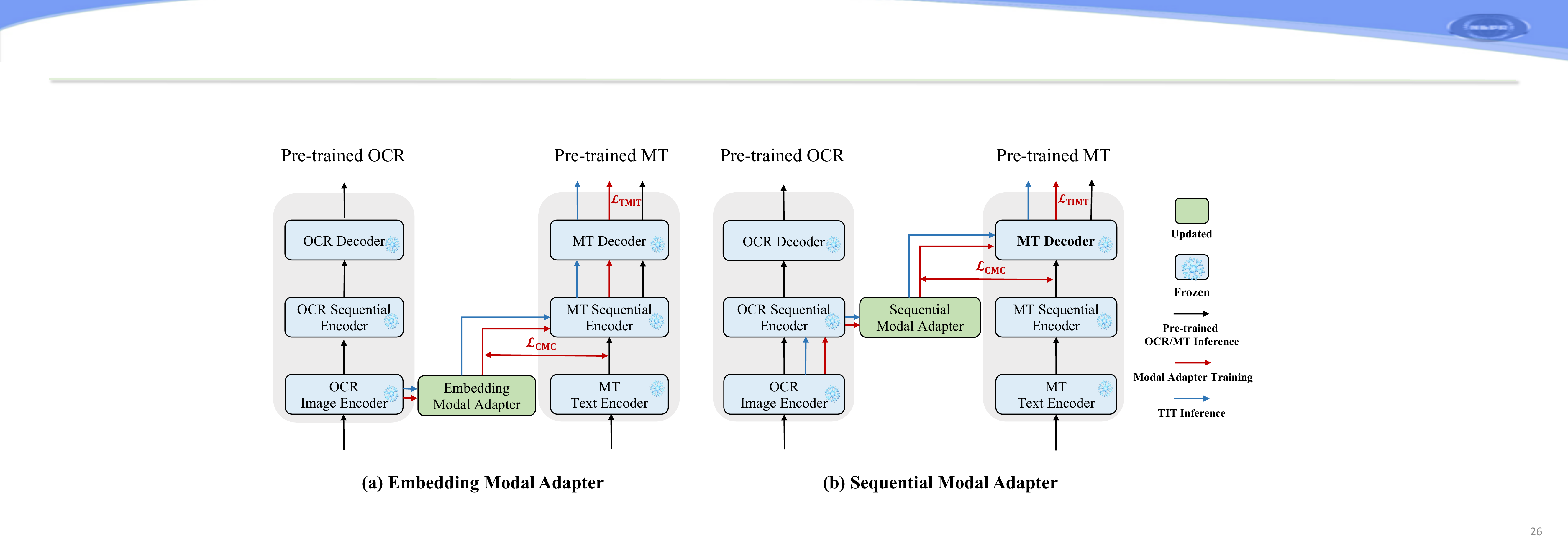}
	\caption{Diagram of (a) Embedding Modal Adapter and (b) Sequential Modal Adapter. Black, red and blue arrow lines denote the pre-trained OCR/MT, modal adapter training and TIMT inference flows respectively. The green box refers to trainable parameters and the blue box refers to frozen ones.}
	\label{fig_seq_ma_arch}       
	\vspace{-0.71cm}
\end{figure*}

Since there are two submodules in the OCR encoder (image encoder and sequential encoder) as introduced in Section 2.1, we propose two types of modal adapters.
The first one is the embedding modal adapter (EmbMA), which aims at aligning the image embedding and text embedding. 
The second one is the sequential modal adapter (SeqMA), which transforms the sequential features encoded by the image sequential encoder to the sequential feature space of the MT task. We will introduce our proposed EmbMA and SeqMA in detail.

\subsection{Embedding Modal Adapter}
The embedding modal adapter is placed in the middle of the OCR image encoder and the text sequential encoder as shown in Figure~\ref{fig_seq_ma_arch} (a).
First, the EmbMA transforms the image embedding into the text embedding space. Second, to better meet the feature distribution of the MT processing flow, the output of EmbMA is constrained by the text embedding through a cross-modal contrastive loss $\mathcal{L}_\text{CMC}^\text{EmbMA}$.
As so, the output of EmbMA given $i$-th image embedding should be similar to the $i$-th text embedding, and apart from the other text embeddings in the mini-batch:

\begin{equation}
\small
\begin{aligned}
H_\text{EmbMA}^{(i)} &= \text{EmbMA}(E_I^{(i)}) \\
\mathcal{L}_\text{CMC}^\text{EmbMA} &= -\sum_{i=1}^{K}\text{log}\frac{\text{exp}(d(H_\text{EmbMA}^{(i)}, E_T^{(i)})/\tau)}{\sum_{j=1}^{K}\text{exp}(d(H_\text{EmbMA}^{(i)}, E_T^{(j)})/\tau)}
\end{aligned}
\end{equation}
where $\text{EmbMA}(\cdot)$ utilizes the same modal adapter architecture as in Equation 7. $H_\text{EmbMA}^{(i)}$ represents the output of the EmbMA. $E_I^{(i)}$ and $E_T^{(i)}$ denote the image and text embedding of $i$-th sample respectively. $K$ denotes the size of the mini-batch. $\tau$ stands for the temperature parameter and $d(q,k)$  represents the similarity metric which we utilize cosine similarity in our implementation.

Aligned with text embedding, the outputs of EmbMA are further fed into the text sequential encoder to obtain the contextual feature $S_\text{EmbMA}^{(i)}$. Through EmbMA, the image embeddings are transformed into the MT processing flow,
and MT decoder finally generates target translation:

\begin{equation}
\begin{aligned}
S_\text{EmbMA}^{(i)} &= \text{Seq}_T(H_\text{EmbMA}^{(i)}) \\
D_\text{EmbMA}^{(i)} &= \text{Dec}_T(S_\text{EmbMA}^{(i)}) \\
P(Y^{(i)}|I^{(i)}) &= \text{Softmax}(W_TD_\text{EmbMA}^{(i)})
\end{aligned}
\end{equation}

\subsection{Sequential Modal Adapter}
Different from EmbMA, SeqMA is designed to align the sequential features of the OCR and MT models. 
As shown in Figure~\ref{fig_seq_ma_arch} (b), SeqMA first transforms the image sequential features into text sequential feature space. Then, the MT decoder generates target language tokens given transformed image sequential features:

\begin{equation}
\begin{aligned}
H_\text{SeqMA}^{(i)} &= \text{SeqMA}(S_I^{(i)}) \\
D_\text{SeqMA}^{(i)} &= \text{Dec}_T(H_\text{SeqMA}^{(i)}) \\
P(Y^{(i)}|I^{(i)}) &= \text{Softmax}(W_TD_\text{SeqMA}^{(i)})
\end{aligned}
\end{equation}
where $\text{SeqMA}(\cdot)$ uses the same structure as in Equation 7. $H_\text{SeqMA}^{(i)}$ denotes the output of the sequential modal adapter and $S_I^{(i)}$ represents the output of the image sequential encoder of the $i$-th sample in the mini-batch. $D_\text{SeqMA}^{(i)}$ denotes the output of text decoder given the hidden states from SeqMA.

Since the hidden states of the SeqMA are further fed into the MT decoder, the feature distribution of $H_\text{SeqMA}^{(i)}$ should be similar to the hidden states of $S_T^{(i)}$. To bridge the feature gap between the OCR and MT tasks, a cross-modal contrastive loss is utilized to align the feature distribution of the transformed image sequential feature and text sequential feature:

\begin{equation}
\small
\begin{aligned}
\mathcal{L}_\text{CMC}^\text{SeqMA}=-\sum_{i=1}^{K}\text{log}\frac{\text{exp}(d(H_\text{SeqMA}^{(i)}, S_T^{(i)})/\tau)}{\sum_{j=1}^{K}\text{exp}(d(H_\text{SeqMA}^{(i)}, S_T^{(j)})/\tau)} \\
\end{aligned}
\end{equation}	
where $d(\cdot)$ and $\tau$ are the same similarity metric and temperature parameter as introduced in Equation 8.

\subsection{Training of Modal Adapter}
During model training, only parameters in modal adapters are updated, while the parameters in the OCR and MT models are all fixed. Through parameter-efficient modal adapter tuning, the pre-trained OCR encoder and MT decoder are able to transfer to the TIMT task with ease. Specifically, multi-task learning is utilized by optimizing end-to-end text image translation loss and cross-modal contrastive loss. Formally, the end-to-end text image translation loss and the overall loss functions are:

\begin{equation}
\begin{aligned}
\mathcal{L}_\text{TIMT}&=-\sum_{i=1}^{|D_\text{TIMT}|} \text{log}P(Y^{(i)}|I^{(i)}) \\
\mathcal{L}_\text{All}&=(1-\lambda_\text{CMC})\mathcal{L}_\text{TIMT} + \lambda_\text{CMC}\mathcal{L}_\text{CMC}
\end{aligned}
\end{equation}
where $\mathcal{L}_\text{CMC}$ is introduced as in Equation 8 and Equation 11. $\lambda_\text{CMC}$ denotes the hyper-parameter, which balances the weight of end-to-end text image translation loss and cross-modal contrastive loss. Note that the $P(Y^{(i)}|I^{(i)})$ in end-to-end text image translation loss $\mathcal{L}_\text{TIMT}$ and cross-modal contrastive loss $\mathcal{L}_\text{CMC}$ are calculated based on the corresponding training workflow of SeqMA and EmbMA.

\subsection{Inference}
During model inference, as the blue arrow lines shown in Figure~\ref{fig_seq_ma_arch}, the input images are first fed into the OCR encoder to obtain the image features. Second, the modal adapter transforms the image features into the MT feature space, and the MT decoder finally generates translation results. Note that the OCR decoder and the MT encoder are not utilized during inference resulting in a fast decoding speed with the end-to-end processing architecture as shown in Figure~\ref{fig_overall_workflow} (c). By bridging OCR encoder and MT decoder, modal adapter based method can take full advantage of pre-trained OCR and MT models.

\section{Experiments}
\subsection{Datasets}
OCR, MT, and end-to-end TIMT datasets are utilized in our experiments. OCR and MT datasets are used to train the OCR and MT models respectively. While the TIMT dataset is used to train the parameters in the modal adapter.

\paragraph{OCR Datasets.} OCR datasets are composed of text images and corresponding text pairs $\{(I_i, T_i)\}_{i=1}^{|D_\text{OCR}|}$. Three OCR datasets are considered in our experiments. \textbf{MJSynth (MJ)}~\cite{DBLP:journals/corr/JaderbergSVZ14}\footnote{https://www.robots.ox.ac.uk/vgg/data/text/} is a synthetic word box image recognition dataset designed for English scene text recognition containing 8.9M synthetic word box images. \textbf{SynthText (ST)}~\cite{DBLP:conf/cvpr/GuptaVZ16}\footnote{https://www.robots.ox.ac.uk/vgg/data/scenetext/} is another synthetic dataset containing 5.5M word box images, which renders the texts onto real-world scene images. \textbf{Synthetic Text Line Dataset} is a customized text line recognition dataset that is constructed with the rule-based synthetic method\footnote{https://github.com/Belval/TextRecognitionDataGenerator}. 1M English and 1M Chinese synthetic text line recognition pairs are synthesized in our experiments.

\paragraph{MT Datasets.} Parallel sentences from the Workshop of Machine Translation 2018~\footnote{http://www.statmt.org/wmt18/} are utilized to train the text machine translation models. Specifically, three translation directions are considered in our experiments: English-to-Chinese (En$\Rightarrow$Zh), English-to-German (En$\Rightarrow$De), and Chinese-to-English (Zh$\Rightarrow$En). After pre-processing and filtering, 5,984,287 En$\Leftrightarrow$Zh and 20,895,771 En$\Rightarrow$De translation pairs are finally obtained to train MT models.

\paragraph{End-to-End TIMT Datasets.} A public end-to-end TIMT dataset proposed by~\cite{DBLP:conf/icpr/MaZTHWZ022} is utilized to train end-to-end TIMT models. This dataset is a synthetic text image translation corpus by synthesizing the text image through a rule-based toolkit given randomly selected background images, font types, and other rendering effects, which is similar to the synthesis method as synthetic text line recognition dataset. The parallel sentences of the end-to-end text image translation datasets are extracted from the text translation corpus. In summary, one million end-to-end TIMT pairs are utilized for each translation direction.

\paragraph{Evaluation Datasets.} Evaluation sets constructed by~\cite{DBLP:conf/icpr/MaZTHWZ022} are used to measure the performance of various models. Three domains are considered, including synthetic, subtitle, and street-view evaluation domains. The synthetic evaluation dataset contains 2,502 En$\Leftrightarrow$Zh and 2,000 En$\Rightarrow$De translation pairs, which are synthesized as the synthetic training dataset. For real-world evaluation datasets, the En$\Leftrightarrow$Zh subtitle dataset contains 1,040 translation pairs, while the En$\Rightarrow$Zh street-view dataset contains 1,198 translation pairs.

\begin{table*}[t]
\centering
\renewcommand{\arraystretch}{1.1}
\setlength{\tabcolsep}{0.9mm}{
\caption{Comparison of end-to-end, cascade and modal adapter tuning based text image machine translation models.}
\label{E2E_vs_Cascaded_vs_MA}
\begin{tabular}{l|lll|ll|l}
\toprule[0.4mm]
 \multirow{2}{*}{Architecture} & \multicolumn{3}{c|}{Synthetic} & \multicolumn{2}{c|}{Subtitle} & Street \\
  & En$\Rightarrow$Zh & En$\Rightarrow$De & Zh$\Rightarrow$En & En$\Rightarrow$Zh & Zh$\Rightarrow$En & Zh$\Rightarrow$En \\
\hline
\multicolumn{7}{c}{End-to-End Models} \\
\hline
 TRBA~\cite{DBLP:conf/iccv/BaekKLPHYOL19} & 9.61 & 7.36 & 4.77 & 12.12 & 5.18 & 0.36 \\
 CLTIR~\cite{DBLP:conf/icpr/ChenYZYL20} & 18.02 & 15.55 & 10.74 & 16.47 & 9.04 & 0.43 \\
 CLTIR+OCR~\cite{DBLP:conf/icpr/ChenYZYL20} & 19.44 & 16.31 & 13.52 & 17.96 & 11.25 & 1.74 \\
 RTNet~\cite{DBLP:conf/icdar/SuLZ21} & 18.91 & 15.82 & 12.54 & 17.63 & 10.63 & 1.07 \\
 RTNet+OCR~\cite{DBLP:conf/icdar/SuLZ21} & 19.63 & 16.78 & 14.01 & 18.82 & 11.50 & 1.93 \\
 \multirow{1}{*}{MTETIMT\cite{DBLP:conf/icpr/MaZTHWZ022}} & 19.25 & 16.27 & 13.16 & 17.73 & 10.79 & 1.69 \\
 MTETIMT+MT\cite{DBLP:conf/icpr/MaZTHWZ022} & {21.96} & {18.84} & {15.62} & {19.17} & {12.11} & {5.84} \\
MHCMM\cite{ChenZhuo:TMM} & 22.08 & 18.97 & 15.66 & 19.24 & 12.12 & 5.87 \\
\hline
\multicolumn{7}{c}{Cascade Models} \\
\hline
 CRNN\ +\ Transformer & 14.43 & 11.27 & 10.52 & 17.88 & 10.06 & 3.25 \\
 TRBA\ +\ Transformer & 17.59 & 13.86 & 12.79 & 18.22 & 10.53 & 4.08 \\
 TRT\ +\ Transformer & 20.46 & 16.48 & 15.12 & 19.12 & 12.08 & 5.78 \\
\hline
\multicolumn{7}{c}{Modal Adapter Tuning Models} \\
\hline
 Sequential Modal Adapter & 20.90 & 19.02 & 15.22 & 19.31 & 12.03 & 5.81 \\
 Embedding Modal Adapter & \textbf{22.53} & \textbf{19.67} & \textbf{16.25} & \textbf{19.46} & \textbf{12.39} & \textbf{6.24} \\
\bottomrule[0.4mm]
\end{tabular}}
\vspace{-0.35cm}
\end{table*}

\vspace{-0.175cm}
\subsection{Experimental Settings}
We implement the image encoder based on the code release by ~\cite{DBLP:conf/iccv/BaekKLPHYOL19}. The MT model is utilized the same architecture proposed in ~\cite{DBLP:conf/nips/VaswaniSPUJGKP17}. The OCR and MT models are firstly trained with OCR and MT datasets respectively. Parameters of OCR and MT models are then frozen during fine-tuning. The implementation of the modal adapter is utilized a similar architecture as the transformer encoder with the hidden dimensions of 512, 8 attention heads, and a dropout rate of 0.1. The initial learning rate is set to 2e-3, the batch size is 64, and the training step is set to 300,000. Parameters of the modal adapter are initialized with Xavier initiation method ~\cite{DBLP:journals/jmlr/GlorotB10} and optimized with Adam optimizer~\cite{DBLP:journals/corr/KingmaB14} on single NVIDIA V100 GPU. Detokenized BLEU~\cite{DBLP:conf/acl/PapineniRWZ02} calculated by sacre-BLEU~\footnote{https://github.com/mjpost/sacrebleu} is utilized as the metric to evaluate the performance of text image translation models.

\vspace{-0.175cm}
\subsection{Comparison of Various Text Image Translation Models}
Table~\ref{E2E_vs_Cascaded_vs_MA} shows the BLEU scores of text image translation models on various evaluation datasets. Three OCR models are utilized in the cascade models: CRNN~\cite{DBLP:journals/pami/ShiBY17}, TPS+ResNet+BiLSTM+Attention (TRBA)~\cite{DBLP:conf/iccv/BaekKLPHYOL19}, and TPS+ResNet+\\Transformer (TRT). While transformer-base~\cite{DBLP:conf/nips/VaswaniSPUJGKP17} is utilized for MT model. The performance of the OCR and the MT models are shown in Section~\ref{ocr_and_mt_performance}.
Five architectures are compared in end-to-end TIMT setting. TRBA~\cite{DBLP:conf/iccv/BaekKLPHYOL19} represents the OCR architecture trained with end-to-end TIMT dataset. CLTIR~\cite{DBLP:conf/icpr/ChenYZYL20} model trains end-to-end TIMT with auxiliary OCR task. RTNet~\cite{DBLP:conf/icdar/SuLZ21} utilizes a feature transformer to link OCR encoder and decoder but ignores the task gap modeling. MTETIMT~\cite{DBLP:conf/icpr/MaZTHWZ022} represents the machine translation enhanced end-to-end TIMT model, which utilizes multi-task learning with auxiliary translation task. While MHCMM\cite{ChenZhuo:TMM} proposes a multi-hierarchy cross-modal mimic framework for the end-to-end text image translation, which incorporates external text translation corpus and utilizes text MT model as teacher guidance for TIMT model.
The modal adapter in Table~\ref{E2E_vs_Cascaded_vs_MA} bridges the pre-trained TRT OCR encoder and transformer MT decoder. 
Experimental results show that our proposed sequential and embedding modal adapter outperforms two-stage cascade models on three translation domains with an average improvement of 1.01 BLEU scores. 
Meanwhile, modal adapter improves the TIMT performance on various language directions (En$\Rightarrow$Zh and En$\Rightarrow$De), revealing the method is robust to different language settings. For Zh$\Rightarrow$En translation direction, modal adapter based method achieves similar results as the previous machine translation enhanced multi-task training model, indicating modal adapter method can take full advantage of the pre-trained MT model without multi-task training.

Furthermore, the embedding modal adapter performs better than the sequential modal adapter, and we attribute that EmbMA retains the cross-attention flow between the original text sequential encoder and decoder.
This shows it is vital not only to eliminate the gap between the OCR and MT tasks but also to maintain the consistency of structures within each task.

\begin{table*}[t]
\centering
\renewcommand{\arraystretch}{1.1}
\setlength{\tabcolsep}{0.5mm}{
\caption{Performance of text image recognition models. Metric of scene text recognition (Rec.) is word accuracy and character error rate is utilized for text line recognition evaluation. Tr.E and Tr.D represent transformer encoder and decoder respectively.}
\label{performance_of_ocr}
\begin{tabular}{l|c|c|c|lll|lll}
\toprule[0.4mm]
\multirow{3}{*}{Architecture} & \multirow{3}{*}{\makecell[c]{Image\\Encoder}} & \multirow{3}{*}{\makecell[c]{Image\\Sequential\\Encoder}} & \multirow{3}{*}{Decoder} & \multicolumn{3}{c|}{Scene Text Rec.} & \multicolumn{3}{c}{Text Line Recognition} \\
\cline{5-10}
& & & & IIIT  & SVT & SP  & Synthetic & Subtitle & Street \\
& & & & 3000 & 647 & 645 & 2502 & 1040 & 1198 \\
\hline
CRNN~\cite{DBLP:journals/pami/ShiBY17} & VGG & BiLSTM & CTC & 81.3 & 79.0 & 66.7 & 13.90 & 4.95 & 56.82 \\
TRBA~\cite{DBLP:conf/iccv/BaekKLPHYOL19} & ResNet & BiLSTM & Attention & 86.6 & 87.8 & 76.9 & 12.29 & 3.01 & 51.67 \\
TRT & ResNet & Tr.E & Tr.D  & 87.9 & 87.2 & 78.6 & 10.89 & 2.33 & 49.83 \\
\bottomrule[0.4mm]
\end{tabular}}

\vspace{-0.15cm}

\end{table*}

\begin{table*}[t]
\centering

\renewcommand{\arraystretch}{1.1}
\setlength{\tabcolsep}{1.6mm}{
\caption{Performance of text translation models. BLEU score is utilized as the metric of text translation task.}
\label{performance_of_mt}
\begin{tabular}{l|lll|ll|l}
\toprule[0.4mm]
\multirow{2}{*}{Architecture} & \multicolumn{3}{c|}{Synthetic} & \multicolumn{2}{c|}{Subtitle} & Street \\
& En$\Rightarrow$Zh & En$\Rightarrow$De & Zh$\Rightarrow$En & En$\Rightarrow$Zh & Zh$\Rightarrow$En & Zh$\Rightarrow$En \\
\hline
Transformer-Base~\cite{DBLP:conf/nips/VaswaniSPUJGKP17} & 25.38 & 20.97 & 17.56 & 19.64 & 13.78 & 15.17 \\
Transformer-Big\ \ ~\cite{DBLP:conf/nips/VaswaniSPUJGKP17} & 26.41 & 22.15 & 19.04 & 20.39 & 14.66 & 16.93 \\

\bottomrule[0.4mm]
\end{tabular}}

\vspace{-0.35cm}

\end{table*}

\vspace{-0.175cm}
\subsection{Performance of OCR and MT Models}
\label{ocr_and_mt_performance}
OCR and MT models in cascade models are firstly trained with corresponding OCR and MT datasets. Parameters in pre-trained OCR and MT models are then frozen during modal adapter training.
Three OCR models CRNN, TRBA, and TRT are all trained with the same scene text recognition and synthetic text line recognition datasets introduced in section 4.1. 
Table~\ref{performance_of_ocr} shows the performance of various OCR models. Transformer based TRT model achieves the best recognition performance, indicating the strong sequential encoder is essential for optical character recognition.
For MT models, the transformer-base and the transformer-big~\cite{DBLP:conf/nips/VaswaniSPUJGKP17} are utilized to translate the source language into the target language.
Table~\ref{performance_of_mt} shows the performance of text translation, and the transformer-big achieves better translation BLEU.

\begin{figure*}[t]
	\centering
	\includegraphics[scale=0.45]{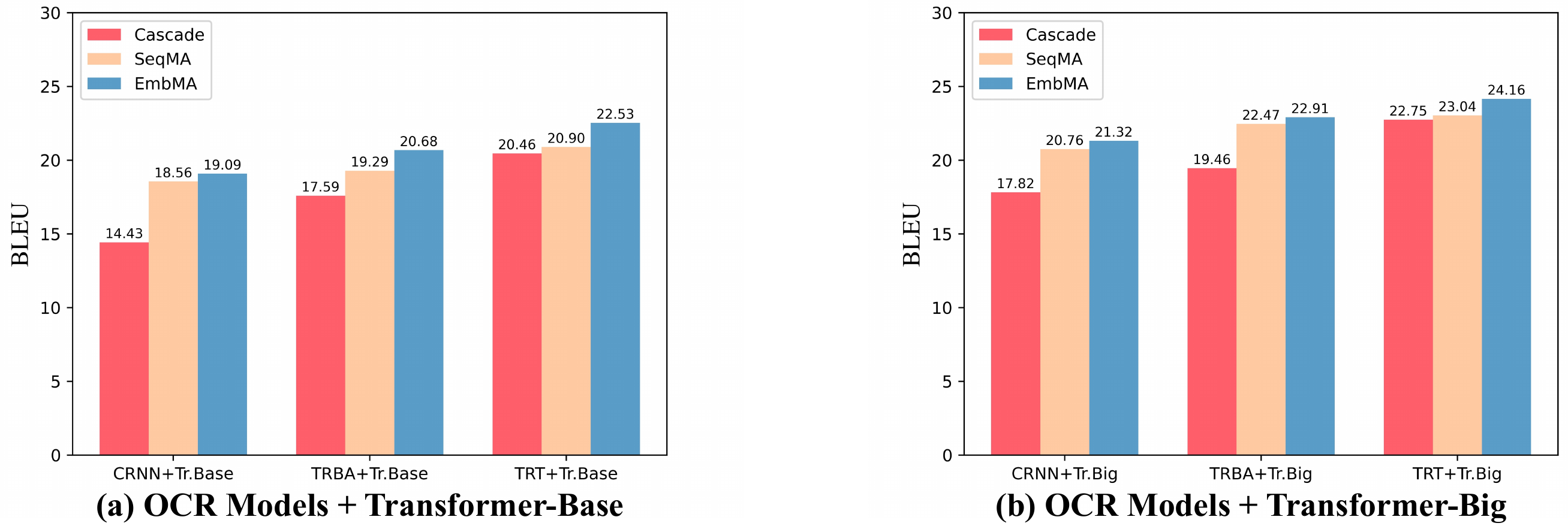}
	\caption{Performance of various OCR and MT combinations with modal adapter. CRNN, TRBA, and TRT represent three OCR models. While MT Models include transformer-base (Tr.Base) and transformer-big (Tr.Big).}
	\label{fig_ocrs_and_trbase}       
\end{figure*}

\vspace{-0.175cm}
\subsection{Generalization of Modal Adapter on Various OCR and MT Combinations}
\label{generalization}
To evaluate the generalization of our proposed method, the modal adapter is studied by bridging various OCR encoders and MT decoders. 
As shown in Figure~\ref{fig_ocrs_and_trbase}, modal adapter tuning outperforms the cascade models on different OCR and MT combinations, revealing the good generalization of modal adapter tuning methods. Figure~\ref{fig_ocrs_and_trbase} (a) shows the text image translation results by combining different OCR models and transformer base MT model. Better OCR image encoder can extract more information into image features, leading to better text image translation performance. 

Figure~\ref{fig_ocrs_and_trbase} (b) depicts various OCR models with transformer big MT models. Similar to Figure~\ref{fig_ocrs_and_trbase} (a), better OCR models achieve better results with transformer big MT models. Furthermore, stronger MT decoders can further improve the translation performance in Figure~\ref{fig_ocrs_and_trbase} (b) compared with Figure~\ref{fig_ocrs_and_trbase} (a).
As a result, our proposed modal adapter tuning method has strong scalability by bridging better OCR and MT models.

\vspace{-0.35cm}
\subsection{Analysis on Model Size and Decoding Speed of TIMT Models}
Cascade models have redundant parameters and slow decoding speed. By removing the OCR decoder and the MT encoder, the modal adapter tuning method has fewer parameters and a faster decoding speed. As shown in Table~\ref{comparison_of_params_speed}, the end-to-end model, which is trained from the scratch, has fewer parameters and a faster decoding speed compared with the cascade model. Fine-tuning model is also an end-to-end model, which is initialized with the OCR encoder and MT decoder. Then the fine-tuning model is trained with the end-to-end text image translation dataset. 
Since the modal adapter bridges the OCR encoder and the MT decoder directly, 
it has a faster decoding speed than the cascade model. Meanwhile, after removing the OCR decoder and MT encoder, modal adapter models have fewer parameters than the cascade model. 
For the comparison of fine-tuning methods, modal adapter tuning outperforms fine-tuning model, because modal adapter models the task consistency between the OCR encoder and MT decoder, which alleviates the gap between OCR and MT tasks.

\begin{table*}[t]
\centering
\renewcommand{\arraystretch}{1.2}
\setlength{\tabcolsep}{0.7mm}{
\caption{Comparison of model size and decoding speed among various models on English-to-Chinese translation direction. The unit of parameters is million ($\times 10^6$), while the unit for speed is sentence per second. BLEU score is utilized to show the performance of synthetic and subtitle text image translation.}
\label{comparison_of_params_speed}
\begin{tabular}{lcllll}
\toprule[0.4mm]
\multirow{1}{*}{Architecture}  & Finetuned Params. & Total Params. & Speed & Synthetic & Subtitle \\
\hline
Cascade & - & 195.1M & 3.07 & 20.46 & 19.12 \\
End-to-End & - & 121.9M ($\downarrow$37.52\%) & 5.21 ($\uparrow$1.70x) & 19.63 & 18.82 \\ 
Fine-tuning & 121.9M & 121.9M ($\downarrow$37.52\%) & 5.21 ($\uparrow$1.70x) & 20.18 & 19.04 \\ 
\hline
\multirow{1}{*}{SeqMA} & \multirow{2}{*}{13.2M} & \multirow{2}{*}{135.1M ($\downarrow$30.75\%)} & \multirow{2}{*}{5.12 ($\uparrow$1.67x)} & 20.90 & 19.31 \\
\multirow{1}{*}{EmbMA} & & & & \textbf{22.53} & \textbf{19.46} \\

\bottomrule[0.4mm]
\end{tabular}}
\end{table*}

\begin{table}[t]
\centering
\renewcommand{\arraystretch}{1.2}
\setlength{\tabcolsep}{10.3mm}{
\caption{Comparison of adapter tuning and modal adapter tuning on English-to-Chinese translation.}
\label{comparison_with_adapter_tuning}
\begin{tabular}{lll}
\toprule[0.4mm]
\multirow{1}{*}{Architecture} & Synthetic & Subtitle \\
\hline
Adapter Tuning  & 16.72 & 15.87 \\
SeqMA (Bottleneck) & 18.25 & 16.80 \\
EmbMA (Bottleneck) & 21.82 & 19.35 \\
\multirow{1}{*}{SeqMA} & 20.90 & 19.31 \\
\multirow{1}{*}{EmbMA} & \textbf{22.53} & \textbf{19.46} \\

\bottomrule[0.4mm]
\end{tabular}}

\vspace{-0.35cm}
\end{table}

\vspace{-0.175cm}
\subsection{Comparison with Adapter Tuning}
Adapter tuning~\cite{DBLP:conf/nips/RebuffiBV17} is an effective parameter-efficient fine-tuning method. Different from adapter tuning, which inserted bottleneck modules inside the pre-trained transformer layers, the modal adapter is designed outside the pre-trained models by bridging the separated OCR encoder and the MT decoder. As shown in Table~\ref{comparison_with_adapter_tuning}, the modal adapter significantly outperforms adapter tuning with 5.81 BLEU for the synthetic domain and 3.59 BLEU for the subtitle domain. To offer a more similar architecture, we also put the bottleneck-based adapter outside the pre-trained models, which is similar to our proposed modal adapter tuning. Bottleneck-based modal adapter tuning also outperforms the vanilla adapter tuning, 
revealing the effectiveness of explicitly modeling the transformation mapping from the OCR feature space to the MT feature space.
Finally, self-attention based modal adapter outperforms the bottleneck-based modal adapter, which we attribute to the strong encoding ability of stacked self-attention layers. 

\begin{figure}[t]
	\centering
	\includegraphics[scale=0.55]{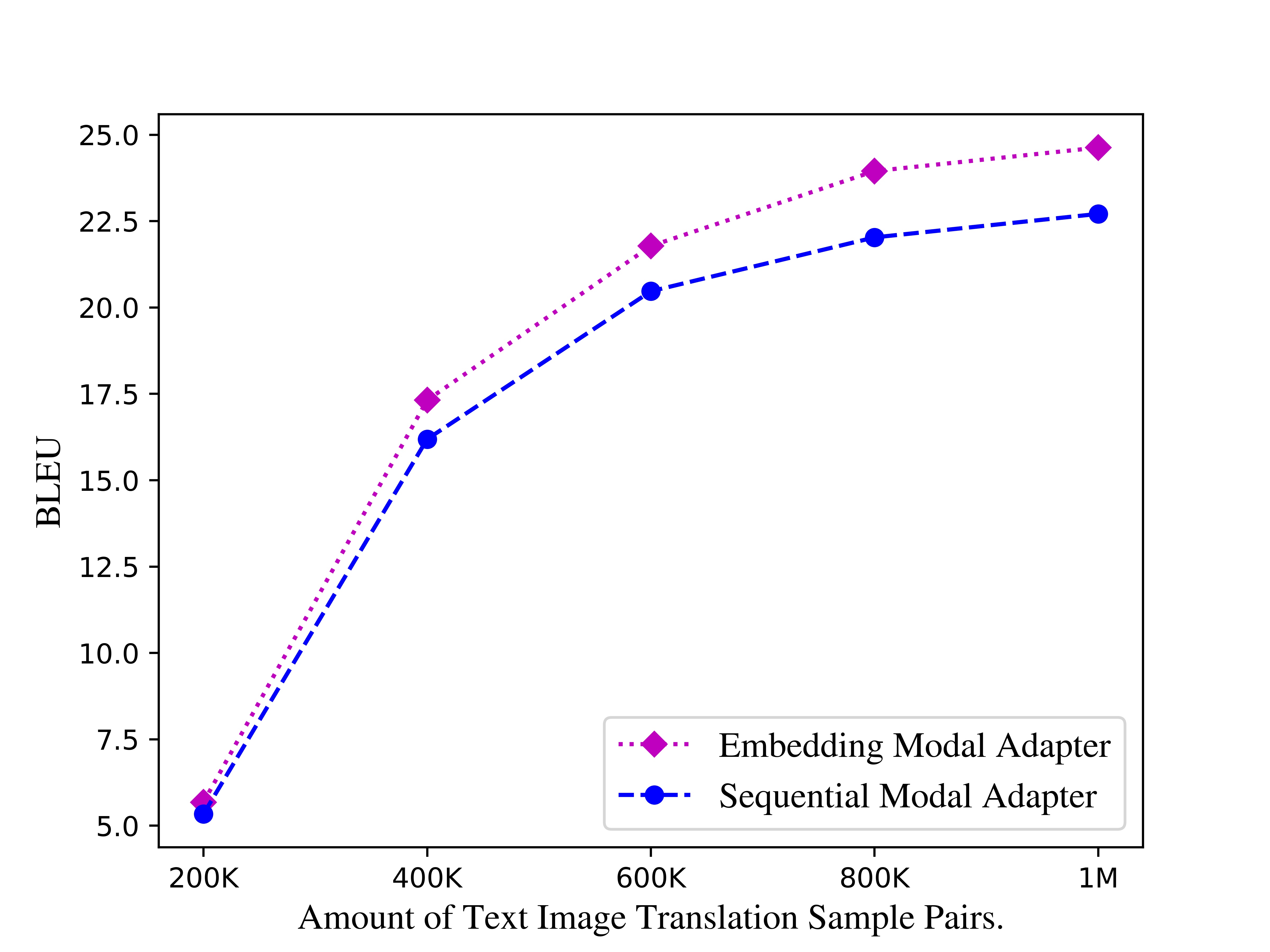}
	\caption{Analysis on the amount of end-to-end TIT datasets on synthetic English-to-Chinese validation set.}
	\label{fig_analysis_on_e2e_data_amount}       
	\vspace{-0.35cm}
\end{figure}

\begin{figure}[t]
	\centering
	\includegraphics[scale=0.55]{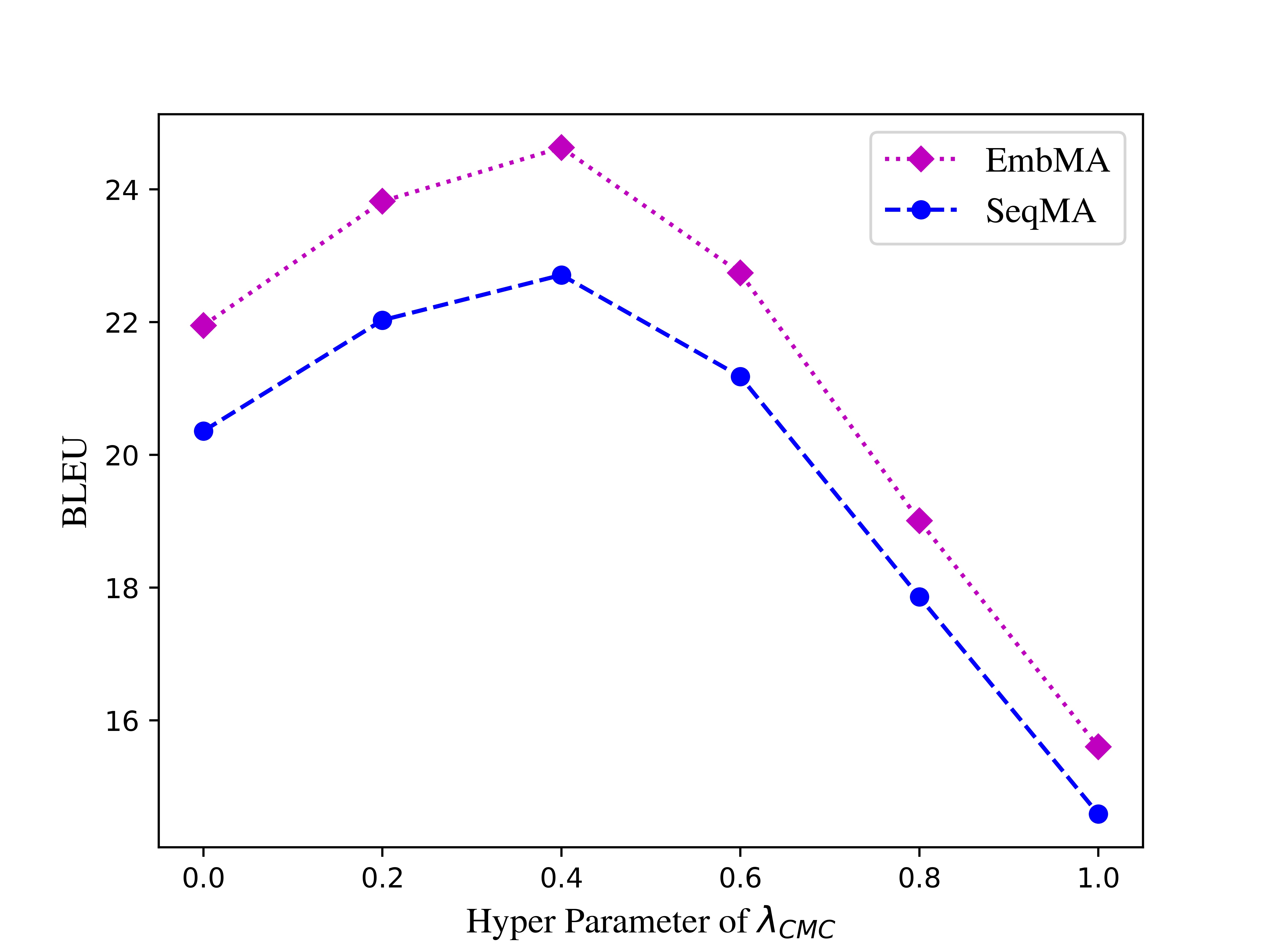}
	\caption{Hyper-parameter evaluation of $\lambda_\text{CMC}$ on English-to-Chinese validation set.}
	\label{fig_hyper_param_analysis}       
	\vspace{-0.35cm}
\end{figure}

\vspace{-0.35cm}
\subsection{Analysis on the Amount of End-to-End TIMT Dataset}
Parameters of the modal adapter are trained on the end-to-end TIMT dataset and the amount of end-to-end data has a great impact on performance.
Figure~\ref{fig_analysis_on_e2e_data_amount} shows the performance of modal adapter tuning with different amounts of end-to-end TIMT datasets. When the end-to-end data is low-resource (around 200 thousand image-text pairs), the performance of modal adapter tuning is limited. We attribute the reason to the non-convergence of modal adapter given low-resource end-to-end data. As the amount of end-to-end TIMT data increases, the modal adapter achieves better results, revealing the modal adapter needs enough data to learn the transformation from the OCR feature space to the MT feature space. When the end-to-end image-text translation data achieves more than 800 thousand pairs, the TIMT results tend to be stable and perform the best translation results. Thus, one million end-to-end text image translation pairs are suitable to train a good end-to-end TIMT model.

\vspace{-0.35cm}
\subsection{Hyper-parameter Analysis}
Hyper-parameter $\lambda_\text{CMC}$ is an important parameter to balance the end-to-end TIMT optimization object and cross-modal contrastive learning object. Figure~\ref{fig_hyper_param_analysis} shows the evaluation of hyper-parameter $\lambda_\text{CMC}$. From this hyper-parameter evaluation, the optimal value of $\lambda_\text{CMC}$ is 0.4 for both embedding modal adapter and sequential modal adapter. When $\lambda_\text{CMC}=0$, parameters in the modal adapter are only optimized by the end-to-end TIMT loss, which ignores the task gap between OCR and MT, leading to performance drop. Specifically, without cross-modal contrastive learning, SeqMA drops 2.35 BLEU scores and EmbMA drops 2.68 BLEU scores, indicating that cross-modal contrastive learning can effectively alleviate the feature gaps between the OCR and MT tasks.  When $\lambda_\text{CMC}=1$, the overall loss function becomes $\mathcal{L}_\text{All}=\mathcal{L}_\text{CMC}$, and the performance drops, indicating end-to-end loss is also vital to modal adapter tuning. Thus, the optimization of the modal adapter should be guided both from direct translation object $\mathcal{L}_\text{TIMT}$ and cross-modal contrastive learning object $\mathcal{L}_\text{CMC}$.

\vspace{-0.35cm}
\section{Related Work}
\vspace{-0.175cm}

\subsection{Text Image Translation.}
TIMT models are mainly divided into the cascade and end-to-end models. Cascade models deploy OCR and MT models respectively~\cite{Shekar2021OpticalCR, manga_translation, DBLP:conf/lt4dh/AfliW16, DBLP:journals/ijdar/ChenCN15, DBLP:conf/icdar/DuHSS11}. Specifically, the source language text images are first fed into OCR models to obtain the recognized source language sentences~\cite{DBLP:conf/cvpr/ShiWLYB16, DBLP:journals/pami/ShiBY17, DBLP:conf/iccv/BaekKLPHYOL19, DBLP:conf/prcv/ZhangNLL19, DBLP:journals/tip/ZhangNLL21, DBLP:journals/soco/KaurK21, DBLP:journals/mta/KaurK21}. Second, the source language sentences are translated into the target language with the MT model~\cite{DBLP:conf/nips/SutskeverVL14, DBLP:conf/nips/VaswaniSPUJGKP17, DBLP:conf/ijcai/ZhaoZZZ20, DBLP:conf/coling/ZhaoXZZZZ20}. Cascade directly connects separated OCR and MT models leading to model redundancy and slow decoding speed. Furthermore, recognition errors made by OCR models are further propagated through MT models, causing severe translation mistakes.

For end-to-end models, the naive approach is to take the OCR model to translate source language text images by training with source language images and corresponding target sentences like TRBA~\cite{DBLP:conf/iccv/BaekKLPHYOL19}. Furthermore, multi-task learning is proposed to incorporate external OCR datasets~\cite{DBLP:conf/icdar/SuLZ21, DBLP:conf/icpr/ChenYZYL20} or MT datasets~\cite{DBLP:conf/icpr/MaZTHWZ022} to enhance the performance of end-to-end models. MHCMM\cite{ChenZhuo:TMM} further improves the feature representation through cross-modal mimic learning on the basis of incorporating external MT data.

However, existing methods still have limitations in fusing cascade and end-to-end models. In this paper, our proposed modal adapter bridges OCR encoder and MT decoder in cascade method through an end-to-end framework, which can take advantage of both cascade and end-to-end methods. Experimental results show modal adapter based TIMT effectively improves translation performance with efficient architecture and fast decoding speed.

\vspace{-0.175cm}
\subsection{Methods of Bridging Encoder and Decoder.} 
Pre-trained models have been explored to achieve good performance after fine-tuning on down-stream tasks~\cite{DBLP:conf/naacl/DevlinCLT19, Radford2018ImprovingLU, DBLP:conf/acl/LewisLGGMLSZ20, DBLP:journals/jmlr/RaffelSRLNMZLL20}. To simplify and speed up the fine-tuning process, efficiency tuning methods are proposed by just updating partial parameters of the model~\cite{DBLP:journals/corr/abs-2211-01642}. Another parameter-efficient tuning research keeps the parameters of pre-trained models unchanged and incorporates external modules to meet the downstream tasks like adapter tuning~\cite{DBLP:conf/nips/RebuffiBV17}, LoRA~\cite{DBLP:conf/iclr/HuSWALWWC22}, BitFit~\cite{DBLP:conf/acl/ZakenGR22}, prefix tuning~\cite{DBLP:conf/acl/LiL20}, and so on. These fine-tuning methods just optimize the parameters of external modules, which makes the fine-tuning process more efficient.

Except for fine-tuning unified pre-trained models, existing research also tried to bridge pre-trained encoder and decoder~\cite{DBLP:journals/tacl/RotheNS20}. ~\cite{DBLP:conf/emnlp/SunW021} proposed to bridge pre-trained mBERT and mGPT through a Graft module to achieve text machine translation. While ~\cite{DBLP:conf/acl/LePWGSB20} explores combining ASR encoder and MT decoder with vanilla adapter for end-to-end speech translation. Inspired by recent research on bridging encoder and decoder, we propose a modal adapter to bridge the OCR encoder and the MT decoder. 

\vspace{-0.175cm}
\section{Conclusion}
\vspace{-0.175cm}
In this paper, we propose a faster and better modal adapter tuning method for the TIMT task, bridging the pre-trained OCR encoder and MT decoder. The sequential modal adapter and embedding adapter are evaluated to verify the effectiveness of bridging different OCR and MT modules. Extensive experiments show embedding modal adapter has better performance because it retains the cross-attention flow between the original MT sequential encoder and decoder. Meanwhile, with an end-to-end architecture, the modal adapter based method outperforms the cascade method with faster decoding speed and lightweight architecture. Furthermore, the modal adapter is effective to bridge various OCR and MT frameworks, revealing the good generalization of the modal adapter tuning method. In the next step, we will design more bridge modules for text image machine translation.

\vspace{-0.175cm}
\section*{Acknowledgement}
\vspace{-0.175cm}
This work has been supported by the National Natural Science Foundation of China (NSFC) grants 62106265.

\bibliographystyle{splncs04}
\bibliography{refs}

\end{document}